\documentclass[letterpaper, 10 pt, conference]{ieeeconf}

\IEEEoverridecommandlockouts

\usepackage{amsmath,amssymb,amsfonts}
\usepackage{graphicx}
\usepackage{textcomp}
\usepackage{newtxtext}
\usepackage{xcolor}
\usepackage{booktabs}
\usepackage{multirow}
\usepackage{float}
\usepackage{cite}
\usepackage[bookmarks=true]{hyperref}
\usepackage{colortbl}
\usepackage{array}
\usepackage{caption}

\begin{document}

\title{\LARGE \bf
ReDex: Repairing Sim-to-Real Dexterous Policies\\
by Finger-Level Compliant Interaction
}

\author{%
\begin{minipage}{\textwidth}
\centering
\fontsize{11}{15}\selectfont
Jinzhou~Li\textsuperscript{1,2}\hspace{0.9em}%
Hadi~Tabatabaee\textsuperscript{1}\hspace{0.9em}%
Kelin~Yu\textsuperscript{1,3}\hspace{0.9em}%
Yuyin~Sun\textsuperscript{1}\hspace{0.9em}%
Cheng-Hao~Kuo\textsuperscript{1}\\[2pt]
Roberto~Mart\'in-Mart\'in\textsuperscript{1,4}\hspace{0.9em}%
Nima~Fazeli\textsuperscript{1,5}\hspace{0.9em}%
X.~Alice~Wu\textsuperscript{1}\hspace{0.9em}%
Xianyi~Cheng\textsuperscript{2}\\[7pt]
{\fontsize{9}{12}\selectfont
\textsuperscript{1}Amazon\hspace{1.5em}%
\textsuperscript{2}Duke University\hspace{1.5em}%
\textsuperscript{3}University of Maryland, College Park\\[2pt]
\textsuperscript{4}The University of Texas at Austin\hspace{1.5em}%
\textsuperscript{5}University of Michigan\endgraf}
\end{minipage}%
}

\maketitle
\begin{abstract}
Dexterous manipulation policies trained in simulation often fail to transfer to the real world because of errors in contact timing and force regulation. Yet these policies can retain useful multi-finger coordination for task progression. We propose ReDex, a framework for adapting a simulation-trained base policy to the real world by correcting
local contact failures and incorporating tactile feedback.
Starting from a proprioception-only base policy, ReDex allows a human operator to physically correct contact failures at selected fingers under compliant control during real-world rollouts, while the frozen base policy continues to control the remaining fingers. These rollouts combine base policy execution, human-corrected finger motion, and fingertip force observations. We reconstruct force-informed targets from these rollouts to train a standalone force-conditioned policy via behavior cloning.
This design reduces human correction effort, enables learning of contact regulation from real-world interaction, and introduces force feedback into a proprioception-only policy without tactile simulation or complex full-hand teleoperation.
We evaluate ReDex on two challenging, contact-rich dexterous manipulation tasks on real hardware. Compared with sim-to-real transferred base policies, ReDex increases Object Flipping success rate from 14\% to 86\% across two objects and average Screwdriver Rotation progress from 26.0\% to 95.3\% across three objects.
\end{abstract}
\section{Introduction}
\label{sec:intro}
For multi-finger hands, large-scale simulation training provides an efficient way to learn the coordination required for dexterous manipulation~\cite{chen2023visual,qi2023hand,wang2024lessons,yang2024anyrotate,kedia2026simtoolreal}. However, sim-to-real transfer proves difficult for several reasons such as errors in contact placement, timing, or contact force regulation. Fingers may touch the wrong surfaces, release contact prematurely, or apply excessive or insufficient force~\cite{liu2025dexndm, wang2024lessons}. Addressing these errors through further simulation training is challenging because real-world contact dynamics are difficult to model accurately. Repeatedly retuning rewards or ranges of domain randomization and retraining policies in simulation can be costly without necessarily resolving the specific contact failures observed in real-world experiments.

\begin{figure}[!t]
  \centering
              \includegraphics[width=0.48\textwidth,trim=0bp 5.5bp 0bp 5.5bp,clip]
    {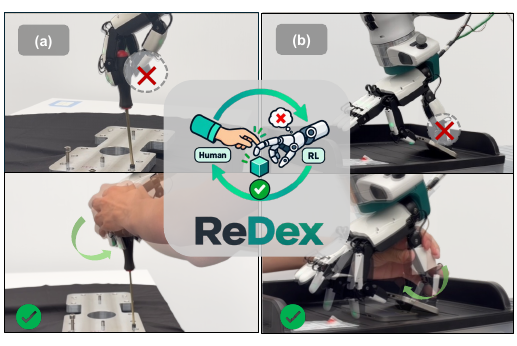}
    \vspace{-5mm}
\caption{\small \textbf{ReDex overview.} ReDex repairs local sim-to-real failures through
finger-level physical correction and reconstruction of training targets. We show \textbf{(a)} Screwdriver Rotation and \textbf{(b)} cellphone Object Flipping. Top: real-world contact failures during base policy execution. Bottom: physical correction of selected fingers. RL denotes the base policy. The corrected rollouts are relabeled with force-informed targets and distilled into a standalone force-conditioned policy via behavior cloning (BC).}
  \label{fig:teaser}
  \vspace{-2em}
\end{figure}

Learning from human corrections on a real robot offers a direct way to address failures due to real-world physics. Existing correction approaches typically rely on teleoperation or kinesthetic teaching to modify the robot’s end-effector or arm-level motion~\cite{jiang2024transic,crdagger}. Extending human corrections to multi-finger hands remains difficult due to three challenges: 1) kinesthetic teaching for multi-finger hands is difficult to execute as finger motions must be simultaneously controlled to ensure stable contact during manipulation, 2) teleoperation does not accurately reflect the optimal contact force profile for completing the target task and 3) collecting successful full-hand demonstrations may require substantial effort.

Existing methods address some of these but not all. A skill-assisted interface reduces this burden by combining human wrist control with learned finger motions~\cite{hsieh2025learning}, but does not allow adjustment of individual finger contacts. Hand-motion retargeting enables finger pose corrections~\cite{han2026dexhil,li2026handitl} but does not ensure appropriate contact forces: a corrected finger pose may still produce insufficient force to prevent slipping or excessive force that impedes motion. Furthermore, identifying and correcting these force errors remains difficult because human operators typically lack direct perception of the contact forces experienced by the robot~\cite{chen2025dexforce, dextac}.
Therefore, a key gap remains in multi-finger dexterous manipulation on enabling human operators to easily correct finger motion and contact forces from simulation-trained policies to enable better task performance and real-world success rate.

To this end, we introduce ReDex, a framework that adapts a simulation-trained dexterous policy to the real world through finger-level physical correction under compliance control, together with real-world fingertip force feedback (Fig.~\ref{fig:teaser}). Our key insight is that these sim-to-real failures are usually \emph{local}: the frozen base policy still produces the multi-finger coordination the task needs, and only a few finger contacts are misplaced or apply the wrong force. ReDex therefore preserves this coordination and lets a human physically repair only the failing fingers, while fingertip tactile sensing captures the contact force each correction applies. Confining correction to where the real world breaks the policy preserves the simulation-learned behavior and keeps the required human effort small.

Specifically, ReDex proceeds in three stages. First, we train the proprioceptive base policy in simulation. Second, we roll out the frozen base policy on the real hand and let the operator physically correct only the fingers that deviate, selectively increasing their compliance so the operator can guide them while the base policy continues to control the remaining fingers in closed loop; each rollout records the full-hand motion and fingertip forces throughout execution, capturing both base-policy-controlled behavior and human physical corrections. Third, we reconstruct force-informed joint-position targets and distill the recorded human corrections and successful base-policy execution into a standalone force-conditioned policy via behavior cloning.

We evaluate ReDex on two challenging dexterous manipulation tasks, Object Flipping and Screwdriver Rotation, across five objects with varying sizes and geometries. Our real-world experiments show that ReDex substantially improves autonomous task performance with a small amount of finger-level correction per object. Ablation studies further highlight the value of tactile feedback both in constructing force-informed supervision and in enabling the learned policy to respond to contact changes during execution.

In summary, our contributions are threefold:
\begin{itemize}
    \item We introduce a human correction framework based on selective finger compliance, enabling direct physical correction of individual finger behavior while the base policy continues to control the remaining fingers, reducing the operator's coordination burden;
    \item We develop efficient learning of finger contact regulation via a relabeling pipeline that turns corrected motion, real-world tactile feedback, and base policy continuation into dense supervision;
    \item We demonstrate that a few minutes of physical correction per object can turn a simulation-trained policy into an improved force-conditioned real-world policy.
\end{itemize}

\begin{figure*}[!t]
\centering

\vspace*{4.5pt}
        \includegraphics[width=0.95\textwidth,trim=0bp 3bp 0bp 2.5bp,clip]
    {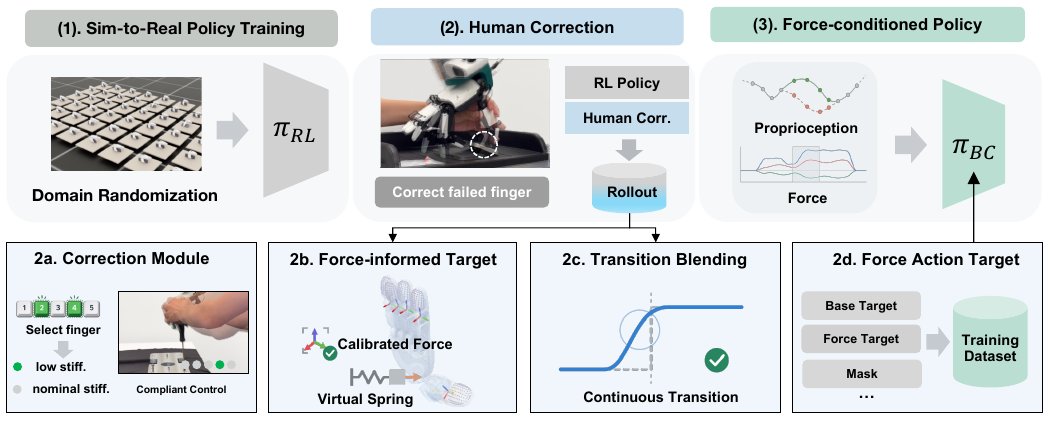}
  \caption{ \small
  \textbf{ReDex pipeline.}
    \textbf{Top:} ReDex follows a three-stage pipeline: \textbf{(1)} train a proprioception-only dexterous policy in simulation, \textbf{(2)} collect real-world finger-level physical corrections through a compliant framework while the base policy controls the remaining fingers, and \textbf{(3)} train a standalone force-conditioned policy from the reconstructed real-world rollouts. \textbf{Bottom:} We detail the correction and supervision-reconstruction process: \textbf{(2a)} finger-level shared control enables selective physical guidance without interrupting the remaining hand motion; \textbf{(2b)} measured fingertip forces are mapped to force-informed joint targets that encode the desired contact load; \textbf{(2c)} these targets are smoothly blended with the evolving base policy targets at intervention onset and release; and \textbf{(2d)} the reconstructed targets form the training supervision for the final policy, which uses proprioception and fingertip force for autonomous deployment.}
  \label{fig:overview}
  \vspace{-1em}
\end{figure*}

\section{Related Work}
\label{sec:related_works}
\subsection{Learning Dexterous Manipulation via RL}
\label{sec:rw-rl}
Training dexterous policies directly on hardware is difficult because collecting demonstrations and running real-world reinforcement learning are costly~\cite{hu2023reboot}. Simulation instead enables large-scale parallel training, privileged state information, and domain randomization for dexterous manipulation, including in-hand reorientation~\cite{qi2023hand,chen2023visual,yang2024anyrotate} and tool use~\cite{kedia2026simtoolreal,hsieh2025learning}. However, sim-to-real transfer remains challenging due to differences in object properties, contact mechanics, and robot actuation, even with domain randomization and system identification~\cite{liu2025dexndm, wang2024lessons, hsieh2025learning}. Our method preserves simulation-learned multi-finger coordination while correcting selected fingers on real hardware.

\subsection{Learning from Human Correction}
\label{sec:rw-hitl}
Human corrections provide a direct way to improve robot policies from deployment failures. Interactive imitation learning methods including DAgger~\cite{ross2011dagger} and its variants~\cite{kelly2019hgdagger,mandlekar2020iwr,liu2023sirius} collect corrective actions on learner-visited states, while human feedback has also been incorporated into real-world reinforcement learning~\cite{luo2025precise}. More recently, human corrections have been used to adapt simulation-trained policies during real-world deployment. TRANSIC~\cite{jiang2024transic} learns a gated residual policy from teleoperated corrections, while CR-DAgger~\cite{crdagger} uses compliant physical interaction to collect corrections and learn residual actions over a base policy. These methods primarily consider arm motion, whereas correction for dexterous hands requires controlling multiple fingers through changing contacts. DexScrew~\cite{hsieh2025learning} reduces this burden by combining human wrist control with a learned full-hand manipulation primitive, preserving learned finger coordination but not allowing direct correction of individual finger contacts. DexHiL~\cite{han2026dexhil} and HandITL~\cite{li2026handitl} instead retarget human hand motion for full-hand or hand--arm intervention, enabling direct finger control but requiring the operator to coordinate the hand during correction. ReDex instead intervenes only on the failed fingers while the base policy continues controlling the remaining fingers, preserving learned coordination during correction.

\subsection{Learning with Contact Feedback}
\label{sec:rw-tactile}
Learning contact-rich manipulation requires both motion coordination and contact regulation. Force feedback provides contact information that is difficult to infer from motion alone. One line of work incorporates real contact information through demonstrations or physical guidance~\cite{chen2025dexforce,kronander2013learning,zhang2025kinedex}. In particular, DexForce~\cite{chen2025dexforce} uses contact forces measured during kinesthetic demonstrations to construct force-informed actions for policy learning. Compliance has also been used to support physical guidance, allowing an operator to directly modify contact through physical interaction~\cite{kronander2013learning,li2014learning,mason1981compliance}. While these approaches provide real contact information, collecting force-rich demonstrations or physical guidance for dexterous manipulation can still require substantial human effort.
Another line of work models tactile sensing in simulation to support contact-aware policy learning and sim-to-real transfer~\cite{dang2026hydroshear,9681378,10912733}. These approaches require sensor-specific modeling and calibration, while accurately simulating contact-rich interactions remains challenging. More recent work introduces force or tactile feedback after learning an initial policy~\cite{hsieh2025learning,clark2026multisensory,crdagger}. We follow this post-training setting and use finger-level compliance to correct only selected fingers while the base policy continues to control the remaining fingers. The measured forces are then used both to reconstruct force-informed supervision and as policy observations, enabling learning from real contact without full-hand force-rich demonstrations.

\section{Method}
\label{sec:method}
Trajectory targets from RL policies can often accomplish dexterous manipulation tasks in simulation, but sim-to-real deployment must account for real-world variations in surface friction, mass/inertia distribution, and hardware degradation over time. Our objective is to correct local finger behavior while retaining the base policy's control of the remaining fingers, and to use these interactions to learn a force-conditioned real-world policy.
As shown in Fig.~\ref{fig:overview}, ReDex adapts a frozen simulation-trained policy (\ref{sec:sim}) through physical correction via finger compliance (\ref{sec:control_framework}), action target reconstruction (\ref{sec:force-target}), and policy learning (\ref{sec:policy}).

\subsection{Sim-to-Real Policy Training}
\label{sec:sim}
Let $q_t\in\mathbb R^n$ denote the measured positions of the hand's $n$ joints. The simulation-trained base policy generates joint position targets $a_t^{\mathrm{RL}}=\pi^{\mathrm{RL}}(\mathbf s_t)$, where $\mathbf s_t$ contains a history of joint positions and previous targets. We train a base policy per task in Isaac Sim~\cite{NVIDIA_Isaac_Sim}, following the Rapid Motor Adaptation (RMA) paradigm~\cite{qi2023hand}. We first train a privileged PPO teacher~\cite{schulman2017proximal} with access to object and dynamics states, then train an adaptation module that estimates the teacher's privileged latent from the history of joint positions and previous targets; the resulting policy $\pi^{\mathrm{RL}}$ requires only proprioception at deployment. \emph{Screwdriver Rotation} uses an object angular-velocity objective~\cite{hsieh2025learning}, whereas \emph{Object Flipping} uses a keypoint-based goal-pose objective~\cite{kedia2026simtoolreal}. To support sim-to-real transfer, we apply task-specific domain randomization to object dynamics, controller gains, and observation and action noise. Each base policy is trained once in simulation and deployed on all corresponding real objects.

\subsection{Finger-Level Compliant Correction}
\label{sec:control_framework}
The key challenge is enabling physical correction without disrupting task execution, since execution and correction require different compliance profiles on the same hand. Reducing stiffness alone is insufficient: a compliant joint is still pulled toward the evolving base policy target, which may conflict with the operator's correction. We combine finger-level compliance control with overrides of the selected joints' position targets. Selected fingers use guidance targets and compliant control settings; unselected fingers track base policy targets under nominal settings.

We realize finger-level compliance through the onboard joint position controller of the RobotEra XHAND1~\cite{roboteraXHand}, which allows independent per-joint position control gains and torque limits. We set the integral gain to zero and use proportional and derivative gains to regulate effective joint stiffness and damping, respectively. Reduced stiffness and torque limits let the operator move selected joints.

At time $t$, the operator selects a finger subset $\mathcal I_t$ using keyboard commands. This selection defines a binary joint-level mask $m_t$, with entries one for joints of selected fingers and zero otherwise. Let $\mathbf k_q^{\mathrm{cmp}}$ and $\mathbf k_q^{\mathrm{nom}}$ denote the position-gain profiles for compliant guidance and nominal policy control,
with $\mathbf k_q^{\mathrm{cmp}}<\mathbf k_q^{\mathrm{nom}}$ elementwise.
The per-joint position gains $\mathbf k_{q,t}$ follow $m_t$:
\begin{equation}
  (k_{q,t})_j =
  \begin{cases}
    (k_q^{\mathrm{cmp}})_j, & (m_t)_j=1,\\
    (k_q^{\mathrm{nom}})_j, & (m_t)_j=0.
  \end{cases}
  \label{eq:compliance-profile}
\end{equation}
Selected joints also use reduced torque limits and a dedicated derivative-gain profile for compliant guidance; unselected joints retain their nominal settings.

When a finger is selected, its guidance target $a_t^h$ is initialized at measured joint positions $q_t$ and then updated to track them, so the position reference follows the measured robot joint configuration. During intervention, the frozen base policy continues to receive the hand's measured joint states, including those of the corrected fingers, and produces full-hand targets $a_t^{\mathrm{RL}}$. The commanded target is
\begin{equation}
  a_t
  =
  m_t\odot a_t^h
  +
  (\mathbf 1-m_t)\odot a_t^{\mathrm{RL}},
  \label{eq:shared-control}
\end{equation}
where $\odot$ denotes elementwise multiplication, so only selected joints receive guidance targets. Upon release, we restore nominal control settings and return these joints to the continuously updated base policy targets. Corrected joint positions remain in the full-hand observation history and can influence base policy outputs for unselected fingers.

During each rollout, we record base policy observations $\mathbf s_t$, measured joint positions $q_t$, base policy joint position targets $a_t^{\mathrm{RL}}$, fingertip forces $\mathbf f_t$, and intervention masks $m_t$. The measured joint positions capture the operator-induced correction motion and, with fingertip forces, yield force-informed joint position targets $a_t^f$ offline.

\subsection{Reconstructing Force-informed Action Targets}
\label{sec:force-target}
Physical correction provides the joint configuration $q_t$, capturing the intended motion but not the desired contact force. We therefore use $q_t$ as the motion component and add an offset from the measured fingertip force. Before each rollout, we tare the fingertip sensors. For each corrected finger $i\in\mathcal I_t$, we threshold measured resultant forces and express each contact force $\mathbf f_{i,t}\in\mathbb R^{3}$ in the hand base frame; $\mathbf f_{i,t}$ is the fingertip force exerted on the object.

Let $\mathbf J_{i,t}\in\mathbb R^{3\times n}$ denote the translational Jacobian of finger $i$'s predefined fingertip frame with respect to all $n$ hand joints at $q_t$. The transpose $\mathbf J_{i,t}^{\top}$ maps fingertip forces to equivalent joint torques:
\begin{equation}
  \boldsymbol{\tau}^{f}_t
  =
  \sum_{i\in\mathcal I_t}
  \mathbf J_{i,t}^{\top}\mathbf f_{i,t}.
  \label{eq:tactile-joint-load}
\end{equation}

Following a quasi-static virtual spring model~\cite{mason1981compliance, chen2025dexforce}, we convert $\boldsymbol\tau^{f}_t$ into joint position
offsets $\Delta a_t^f$ through a fixed diagonal virtual joint stiffness matrix
$\mathbf K_f\in\mathbb R^{n\times n}$ and add them to measured joint
positions $q_t$:
\begin{equation}
  \begin{aligned}
    \Delta a_t^f
    &=\mathbf K_f^{-1}\boldsymbol{\tau}^{f}_t,\\
    a_t^f
    &=q_t+\Delta a_t^f.
  \end{aligned}
  \label{eq:virtual-target}
\end{equation}
Because $\mathbf K_f$ is diagonal and finger joint sets are disjoint, the force-derived offset remains local to corrected fingers.
We set $\mathbf K_f=\gamma^{-1}\operatorname{diag}(\mathbf k^{\mathrm{ref}})$, where
$\mathbf k^{\mathrm{ref}}$ is a reference virtual joint stiffness vector with entries in $\mathrm{N\,m\,rad^{-1}}$, specifying the relative stiffness across joints, and $\gamma$ is a dimensionless scale factor. For fixed $\mathbf k^{\mathrm{ref}}$ and $\boldsymbol\tau_t^f$, the reconstructed joint position offsets scale linearly with $\gamma$. Through preliminary replay experiments, we select $\gamma=0.05$ based on observed contact maintenance and motion
fidelity, and keep it fixed for constructing all policy training targets.

For finger $i$, let $\Delta a_{i,t}^f=\mathbf K_f^{-1}\mathbf J_{i,t}^{\top}\mathbf f_{i,t}$ denote its force-derived offset in the full-hand joint space.
The linearized virtual Cartesian displacement of finger $i$ induced by its measured contact force is
\begin{equation}
\Delta\mathbf x_{i,t}^f
=\mathbf J_{i,t}\Delta a_{i,t}^f
=\underbrace{
\mathbf J_{i,t}\mathbf K_f^{-1}\mathbf J_{i,t}^{\top}
}_{\mathbf C_{x,i,t}}
\mathbf f_{i,t},
\label{eq:force-cartesian-offset}
\end{equation}
where $\mathbf C_{x,i,t}$ is the local virtual Cartesian compliance induced by $\mathbf K_f$ at the current configuration. Thus, $q_t$ captures corrected motion and $\Delta a_t^f$ encodes measured fingertip--object forces in joint targets used only for offline supervision.

We then combine the reconstructed force-informed targets $a_t^f$ and recorded base policy targets $a_t^{\mathrm{RL}}$ into training labels $a_t^\star$ for the complete rollout. During intervention, the joint-level mask $m_t$ identifies corrected joints, whose targets are replaced by the corresponding components of $a_t^f$; unselected joints retain their base policy targets. Force-informed reconstruction is applied only where operator guidance replaces the base policy command. Because the base policy remains in closed loop during data collection, its post-intervention targets reflect task continuation from the corrected state. Switching between $a_t^{\mathrm{RL}}$ and $a_t^f$ can cause abrupt jumps. We smooth training labels $a_t^\star$ offline using minimum-jerk interpolation around intervention boundaries.

\begin{figure*}[t]
\centering
\vspace*{4.5pt}
    \includegraphics[width=0.91\textwidth,trim=0bp 3bp 0bp 1.2bp,clip]
    {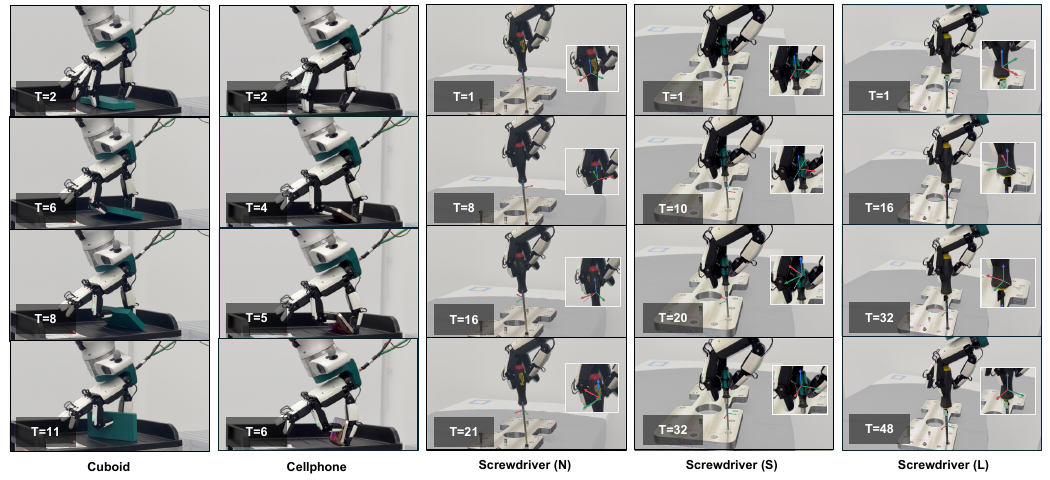}
    \vspace{-1mm}
    \caption{\small \textbf{Qualitative rollout.} Each column shows four frames of autonomous execution after learning from human corrections, ordered from top to bottom. From left to right, we show Object Flipping with a cuboid and a cellphone, and Screwdriver Rotation with normal (N), small (S), and large (L) screwdrivers. Insets visualize changes in screwdriver orientation. T denotes the elapsed time, in seconds, from the start of the corresponding rollout.}
    \label{fig:qualitative-rollouts}
    \vspace{-1mm}
\end{figure*}

\subsection{Learning Force-Conditioned Policy}
\label{sec:policy}
Using the training data $a_t^\star$ from Sec.~\ref{sec:force-target}, we distill successful rollouts into a standalone force-conditioned policy. We train an action-chunking behavior cloning policy~\cite{zhao2023learning} $\pi_\theta$ that maps an observation $x_t=(q_{t-H+1:t},\,\mathbf f_{t-H+1:t})$, containing joint position and fingertip-force histories over $H=5$ steps, to a full-hand joint position chunk of length $N=16$, $A_t^\star=[a_t^\star,\dots,a_{t+N-1}^\star]$. Following standard behavior cloning~\cite{pomerleau1988alvinn}, we optimize
\begin{equation}
    \mathcal{L}_{\mathrm{BC}}(\theta)
    =
    \mathbb{E}_{(x_t,A_t^\star)\sim\mathcal{D}}
    \left[
        \left\|
        \pi_\theta(x_t)-A_t^\star
        \right\|_F^2
    \right],
    \label{eq:bc-objective}
\end{equation}
where $\mathcal D$ denotes the relabeled rollouts. At deployment, the policy predicts joint position chunks from
joint position and fingertip-force histories. We execute the first
12 targets of each 16-step chunk at 20~Hz before predicting
the next chunk.

\section{Experiments}
\label{sec:experiments}
We evaluate our framework in terms of autonomous real-world performance, its use of tactile information and force-informed supervision, and the human effort required to correct. The experiments cover two in-hand manipulation skills and five real objects, with component and representation studies on the cuboid flipping task.
We ask:
\begin{enumerate}
    \item Can ReDex improve the real-world performance of simulation-trained dexterous policies?

    \item How do tactile observations, force-informed targets, and transition blending contribute to real-world adaptation performance?

    \item What level of human effort does ReDex require for real-world correction?
\end{enumerate}

\subsection{Experimental Setup}
\label{sec:exp_setup}
\noindent\textbf{Hardware and Sensing.}
All experiments use the RobotEra XHAND1~\cite{roboteraXHand}, a five-finger, 12-DoF hand with finger compliance control
implemented as described in Sec.~\ref{sec:control_framework}. Each fingertip contains a magnetic tactile array with 120 taxels, each taxel providing a three-axis force measurement. During each rollout, we record taxel measurements and compute each fingertip's three-axis net contact force. We
transform the net forces into the hand-base frame for force-informed action target reconstruction (Sec. ~\ref{sec:force-target}).

\vspace{1mm}
\noindent\textbf{Tasks and metrics.} We evaluate two skills on five real objects. Object Flipping tests brief contact transitions during in-hand reorientation using a cuboid (15 × 7 × 3 cm) and a cellphone (13.15 × 6.42 × 0.74 cm), with dimensions reported as length × width × height. Success requires a $90^\circ$ reorientation ending in a stable grasp.
Screwdriver Rotation tests finger gaiting and sustained contact using normal, small, and large screwdrivers, with handle diameters of 2.6, 1.8, and 3.5 cm, respectively. We measure final progress toward a $360^\circ$ rotation in $90^\circ$ increments, normalized as a percentage. This metric captures partial task completion; we report success rates only for Object Flipping.
The main evaluation uses 25 autonomous trials per object--method pair. The overall flipping result pools 50 trials across both objects, while overall screwdriver progress averages the three object-specific means. An overall value is reported only when all objects in that task family are evaluated. Component and representation variants use 10 trials each; their full-method results are separate from the main 25-trial evaluation.
\newcommand{\nt}{\textcolor{black!45}{\textemdash}}

\begin{table*}[t]
\centering
\vspace*{6pt}
\caption{\small  \textbf{Real-robot performance over 25 autonomous trials per object–method pair.}
Prog. is mean discretized rotation progress, quantized in $90^\circ$
increments and reported as a percentage; SR is the number of successful
trials. Screwdriver Overall is the mean across the three object variants,
while Flipping Overall pools trials across the two object variants.
Overall results are reported only when all corresponding object variants
were evaluated.
}
\label{tab:main}

\footnotesize
\renewcommand{\arraystretch}{1.25}
\setlength{\tabcolsep}{14pt}

\begin{tabular}{lccccccc}
\toprule
& \multicolumn{4}{c}{\textbf{Screwdriver Rotation} (progress.\ \%)}
& \multicolumn{3}{c}{\textbf{Object Flipping} (successes / trials)} \\
\cmidrule(lr){2-5}
\cmidrule(lr){6-8}

Method
& Normal
& Small
& Large
& \textbf{Overall}
& Cuboid
& iPhone
& \textbf{Overall} \\
\midrule

RMA~\cite{qi2023hand}
& 26
& 33
& 19
& 26.0
& 4/25
& 3/25
& 7/50 \\

Gated Corr.~\cite{jiang2024transic}
& 42
& \nt
& \nt
& \nt
& 3/25
& \nt
& \nt \\

Continuous Corr.~\cite{crdagger}
& 36
& \nt
& \nt
& \nt
& 1/25
& \nt
& \nt \\

\rowcolor{gray!12}
\textbf{ReDex (Ours)}
& \textbf{95}
& \textbf{99}
& \textbf{92}
& \textbf{95.3}
& \textbf{20/25}
& \textbf{23/25}
& \textbf{43/50} \\

\bottomrule
\end{tabular}
\end{table*}

\vspace{1mm}
\noindent\textbf{Training setup and details.}
For each skill, we use one frozen simulation-trained policy across objects without additional simulation training and train a separate policy for each object using real-world correction data. We use 30 training episodes per flipping object and 10 per screwdriver variant, selected from successful human-corrected executions.
We train the behavior cloning (BC) policy with separate two-layer MLP encoders for joint positions and force observations. Each encoder has hidden dimensions 128 and 64. We train for 100 epochs using the AdamW optimizer~\cite{loshchilov2017decoupled} with a learning rate of $2\times10^{-4}$ and batch size 64. Inputs and targets are normalized. All methods command joint positions at 20~Hz.

\noindent\textbf{Baselines and comparison scope.}
We evaluate the baselines on cuboid flipping and normal-screwdriver rotation.
RMA directly executes the frozen simulation-trained policy without training on correction data. Gated Correction~\cite{jiang2024transic} and Continuous Correction~\cite{crdagger} learn a residual action that modifies the base policy output. Gated Correction uses a learned binary gate to activate the residual and otherwise executes the base policy alone, whereas Continuous Correction applies a residual at every step.
Both correction baselines use the same rollouts and reconstructed training targets as ReDex.
Their inputs are the base policy joint targets and the same five-frame histories of joint positions and fingertip forces used by ReDex.
Since the base policy produces single-step actions, both baselines predict a single-step residual at each control step. Training targets are expressed as residuals relative to base policy targets. Training and residual combination follow the original methods. These baselines provide reference comparisons with residual-correction approaches, rather than controlled ReDex ablations.
\begin{table}[t]
\centering
\vspace*{6pt}
\caption{\small \textbf{Component ablation on real-robot Object Flipping (Cuboid).} All variants use motion supervision. Checkmarks denote enabled components; SR gives successes over 10 matched trials.}
\label{tab:ablation}
\footnotesize
\renewcommand{\arraystretch}{1.15}
\setlength{\tabcolsep}{3pt}
\begin{tabular}{
>{\raggedright\arraybackslash}p{\dimexpr0.30\columnwidth-2\tabcolsep\relax}
>{\centering\arraybackslash}p{\dimexpr0.12\columnwidth-2\tabcolsep\relax}
>{\centering\arraybackslash}p{\dimexpr0.13\columnwidth-2\tabcolsep\relax}
>{\centering\arraybackslash}p{\dimexpr0.18\columnwidth-2\tabcolsep\relax}
>{\centering\arraybackslash}p{\dimexpr0.12\columnwidth-2\tabcolsep\relax}
}
\toprule
Variant & \shortstack{Force\\obs.} & \shortstack{Virtual\\target} & \shortstack{Transition\\blending} & SR. \\
\midrule
Motion only & -- & -- & -- & 4/10 \\
w Tactile obs. & $\checkmark$ & -- & -- & 6/10 \\
w/o virtual target & $\checkmark$ & -- & $\checkmark$ & 7/10 \\
w/o blending & $\checkmark$ & $\checkmark$ & -- & 6/10 \\
\rowcolor{gray!12}
\textbf{Ours} & $\boldsymbol{\checkmark}$ & $\boldsymbol{\checkmark}$ & $\boldsymbol{\checkmark}$ & \textbf{9/10} \\
\bottomrule
\end{tabular}
\vspace{-1mm}
\end{table}

\section{Results}
\label{sec:exp_main}
\noindent\textbf{Real-world performance.}
We first evaluate ReDex on the cuboid and normal screwdriver, of which the geometry and size are similar to the objects used in simulation training. As shown in Table~\ref{tab:main}, compared to the frozen RMA policy, ReDex increases cuboid success from 4/25 (16\%) to 20/25 (80\%) and normal-screwdriver mean rotation progress from 26\% to 95\%. ReDex also outperforms both correction baselines on these two objects. These results show that our framework corrects failures caused by the base policy, improving real-world performance on both manipulation tasks.

\vspace{1mm}
\noindent\textbf{Adaptation across object variants.}
We next test whether these simulation-trained base policies can be reused through human correction on new objects of different sizes and geometries requiring the same manipulation skill. For each skill, we use one simulation-trained base policy for all objects and train object-specific policies from correction rollouts. As shown in Table~\ref{tab:main}, mean rotation progress improves from 33\% to 99\% on the small screwdriver and from 19\% to 92\% on the large screwdriver, while cellphone flipping success increases from 12\% to 92\%. Together, these results show consistent gains across all objects with limited correction data. We attribute this to the complementary roles of simulation training, domain randomization, and human correction. Simulation training provides shared task progression and multi-finger coordination, domain randomization gives the base policy adaptability to object variation, and the human operator can address remaining local real-world errors. ReDex thus adapts a simulation-trained skill across object variants with limited real-world correction and no further simulation tuning. Fig.~\ref{fig:qualitative-rollouts} shows autonomous flipping and rotation across objects.

\begin{table}[t]
\centering
\vspace*{6pt}
\caption{\small \textbf{Ablation of tactile observation, representation, and encoder on real-robot Object Flipping (Cuboid).} All variants follow a matched 10-trial protocol. Net force stacks one fingertip triaxial resultant per fingertip.}
\label{tab:tac-ablation}
\footnotesize
\renewcommand{\arraystretch}{1.15}
\begin{tabular}{
>{\raggedright\arraybackslash}p{\dimexpr0.85\columnwidth/3-2\tabcolsep\relax}
>{\centering\arraybackslash}p{\dimexpr0.85\columnwidth/3-2\tabcolsep\relax}
>{\centering\arraybackslash}p{\dimexpr0.85\columnwidth/3-2\tabcolsep\relax}
}
\toprule
Tactile Obs. & Encoder & Flip SR. \\
\midrule
Taxels & MLP & 6/10 \\
Taxels & GRU & 2/10 \\
Net force & GRU & 8/10 \\
\rowcolor{gray!12}
\textbf{Net force} & \textbf{MLP} & \textbf{9/10} \\
\bottomrule
\end{tabular}
\vspace{-2mm}
\end{table}

\vspace{1mm}
\noindent\textbf{Comparison with Correction Baselines.}
As shown in Table~\ref{tab:main}, the two correction baselines, Gated Correction~\cite{jiang2024transic} and Continuous Correction~\cite{crdagger}, increase mean rotation progress on the normal screwdriver from RMA's 26\% to 42\% and 36\%, respectively, but achieve only 12\% and 4\% cuboid success, below RMA's 16\%. Our method achieves 95\% rotation progress and 80\% cuboid success. Both correction baselines remain below half of the target rotation on average and fail most flipping trials, indicating limited effectiveness in resolving real-world failures.

We further analyze the behaviors of RMA and the two correction baselines in real-robot experiments. RMA retains coordinated finger motions, but mistimed release can misalign the object
during flipping, while inaccurate finger placement or insufficient contact force can destabilize the screwdriver during rotation. Gated Correction rarely activates despite increased correction weights, while Continuous Correction can disturb useful base motions or fail to counteract erroneous commands. We attribute these difficulties partly to limited correction data and sparse interventions: corrections must be timely and override inappropriate base actions when necessary. These residual baselines also require task-specific tuning of the residual action scale. These observations motivate us to distill human corrections and successful base policy continuations into a standalone policy. ReDex jointly learns local contact repairs and subsequent coordinated motion, retaining the imperfect base policy's skill structure without a gate or residual compensation at deployment.

\begin{figure}[!t]
  \centering
  \vspace*{6pt}
\includegraphics[width=0.43\textwidth,trim=0bp 5.5bp 0bp 0.5bp,clip]{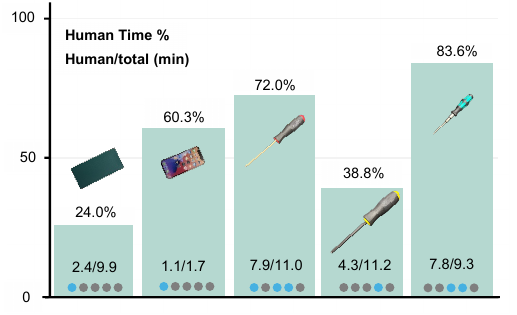}
      \caption{\small \textbf{Human correction across objects.}
  Bars show the share of recorded rollout time with at least one finger
  under human correction. Values inside the bars give human / total rollout
  time in minutes, rounded to 0.1 min. The five circles represent the thumb,
  index, middle, ring, and
  little fingers from left to right; blue marks guided fingers and gray
  marks fingers controlled by the base policy.}
  \label{fig:human-effort}
  \vspace{-1em}
\end{figure}

\vspace{1mm}
\noindent\textbf{Ablation Study.}
We conduct an ablation study on Cuboid Flipping to examine how each component supports learning from human corrections and autonomous execution (Table~\ref{tab:ablation}). Motion supervision alone achieves 4/10 successes, increasing to 6/10 with tactile observations. Without tactile feedback, the policy often stalls during reorientation; with feedback, it adjusts thumb configuration and applied force as contact states change. The complete method achieves 9/10 successes. Removing the force-informed target reduces success to 7/10; the policy sometimes unloads the thumb prematurely, destabilizing supporting contact and causing the object to slip or drop. Removing transition blending reduces success to 6/10. Discontinuities between corrected segments and base policy continuations can cause abrupt finger motions and destabilize contact, even when the correction is appropriate. These results indicate that motion supervision alone does not fully exploit information in human corrections. Human corrections guide manipulation, while tactile observations supply contact-state information, force-informed supervision preserves required contact force information, and transition blending smoothly connects corrections with subsequent actions. Together, these components improve the learned policy's reliability during autonomous execution.

\vspace{1mm}
\noindent\textbf{Tactile representations.}
In addition, we evaluate tactile representations with different encoder architectures. As shown in Table~\ref{tab:tac-ablation}, net-force observations outperform raw taxels with both MLP and GRU encoders. MLP and GRU perform similarly with net-force input, suggesting that this compact representation does not require a complex encoder. Although raw taxels retain richer spatial contact information, their lower performance may reflect greater sensitivity to sensor noise and contact variation. We therefore use net-force observations with an MLP encoder in our experiments.

\vspace{1mm}
\noindent \textbf{Analysis of human effort.}
We analyze operator involvement within retained rollouts using guidance duration and the fingers guided (Fig.~\ref{fig:human-effort}). For Object Flipping (cuboid), we compare 30 trajectories from each method, all validated by replay without human intervention (Table~\ref{tab:human-effort}). ReDex uses 2.4 minutes of guidance applied only to the thumb, compared with 13.2 minutes for full kinesthetic teaching, corresponding to an 82\% reduction in human time. Full kinesthetic teaching also requires guiding both the thumb and index finger. In ReDex, a slight thumb retraction releases contact to allow flipping, after which the base policy continues the task. Thus, human guidance occupies only 24.0\% of retained rollout time, with the base policy handling the rest. The operator corrects a local contact transition rather than the full trajectory; the policy coordinates other fingers.

For the normal, large, and small screwdrivers, guidance occupies 72.0\%, 38.8\%, and 83.6\% of recorded rollout time and involves three, one, and two fingers, respectively (Fig.~\ref{fig:human-effort}). Blue dots mark fingers guided at least once. Screwdriver guidance mainly involves small contact adjustments to the ring finger, with other fingers guided only occasionally. Even during sustained guidance, the base policy continues to control the uncorrected fingers. During the full kinesthetic teaching attempts on Screwdriver Rotation, coordinating repeated contact establishment, release, and reestablishment across fingers was difficult: adjusting one finger could disturb other contacts, causing slipping or interrupting rotation. Even demonstrations that succeeded during teaching could replay inefficiently or unreliably, so we limit the quantitative comparison with full kinesthetic teaching to Cuboid Flipping.

\begin{table}[t]
\centering
\vspace*{6pt}
\caption{\small \textbf{Human guidance effort on Object Flipping
(cuboid).}
Both methods yield 30 trajectories that successfully replay
without human intervention. Reported rollout durations sum the recorded
execution time of these trajectories; they exclude failed attempts, resets,
and replay validation. Both methods use force-informed joint position
targets for replay.}
\label{tab:human-effort}
\footnotesize
\renewcommand{\arraystretch}{1.13}
\setlength{\tabcolsep}{3pt}

\begin{tabular*}{0.9\columnwidth}
{@{\extracolsep{\fill}}lcc@{}}
\toprule
& \shortstack{Full kinesthetic teaching} & \textbf{Ours} \\
\midrule
Guided fingers
    & Thumb, index & Thumb \\
Replay-validated trajectories
    & 30 & 30 \\
\addlinespace[3pt]
Recorded rollout duration (min)
    & 13.2 & 9.9 \\
Human time (min)
    & 13.2 & 2.4 \\
Mean duration per trajectory (s)
    & 26.4 & 19.8 \\
\bottomrule
\end{tabular*}
\vspace{-1em}
\end{table}

\vspace{1mm}
\noindent\textbf{Failure analysis and limitations.}
In our experiments, 1) we observe that the base policy generally maintains useful coordination during local finger corrections. We hypothesize that this behavior reflects both the robustness acquired through simulation training with domain randomization and the restoration of task-relevant contacts through human guidance. However, this tolerance is still limited: excessive or poorly timed interventions can disrupt inter-finger coordination and move the hand outside the policy’s training state distribution. For example, in Screwdriver Rotation, moving the thumb too far from its intended contact point can disrupt the gait phases of other fingers, making subsequent base policy continuation unsuitable for supervision; 2) Distillation from limited real-world trajectories may not fully preserve the generalization ability of the simulation-trained policy, leaving the distilled policy more sensitive to conditions outside the collected data distribution; 3) Magnetic sensor drift remains a limitation, potentially degrading force-derived training labels and policy robustness. Initial taring removes zero offset, but subsequent drift cannot be distinguished from contact force changes without additional references or constraints. 4) Our evaluation is limited to two in-hand skills and a single hardware platform. Extending it to more complex tasks requires further validation of how joint coupling, feedback mechanisms, and intervention magnitude affect coordination and stability.

\section{Conclusion}
ReDex adapts simulation-trained policies through finger-level physical correction and real-world tactile feedback. During rollout, the base policy controls uncorrected fingers while the operator repairs local contact failures. Force-informed target reconstruction and transition blending convert corrected rollouts into supervision for a standalone policy. Across two skills and five objects, ReDex improves real-world performance with limited correction data, without tactile simulation or additional simulation training. Future work could test other tactile sensors, extend ReDex to whole-body manipulation, and investigate preserving simulation-learned adaptability during real-world correction training.

\section*{Acknowledgments}
We thank Haozhi Qi for early discussions about the project, and Elvis Hsieh, Jiahe Pan, Qiyang Yan, and Zeyuan Chen for helpful discussions and suggestions on RL-based primitive skill training. We also thank Qianzhong Chen, Zheda Mai, Zixing Wang, Zhengjie Xu, and Zichao Hu for insightful discussions and feedback, Muyuan Lin for proofreading, and Alvin Ho for his help with 3D printing.

\bibliographystyle{IEEEtran}
\bibliography{reference}

\end{document}